\documentclass[letterpaper]{article} 
\usepackage{aaai23}  
\usepackage{times}  
\usepackage{helvet}  
\usepackage{courier}  
\usepackage[hyphens]{url}  
\usepackage{graphicx} 
\usepackage{natbib}  
\usepackage{caption} 
\usepackage{algorithm}
\usepackage{algorithmic}

\usepackage{amsfonts}
\usepackage{amsmath}
\usepackage{booktabs}
\usepackage{multirow}

\usepackage[resetlabels]{multibib}
\newcites{new}{References}
\usepackage{newfloat}
\usepackage{listings}
\DeclareCaptionStyle{ruled}{labelfont=normalfont,labelsep=colon,strut=off} 
\floatstyle{ruled}
\newfloat{listing}{tb}{lst}{}
\floatname{listing}{Listing}
\title{PIXEL: Physics-Informed Cell Representations\\for Fast and Accurate PDE Solvers}
\author{
    Namgyu Kang\textsuperscript{\rm 1},\;\, Byeonghyeon Lee\textsuperscript{\rm 1},\;\, Youngjoon Hong\textsuperscript{\rm 2},\;\, Seok-Bae Yun\textsuperscript{\rm 2},\;\, Eunbyung Park\textsuperscript{\rm 1, \rm 3}\thanks{Corresponding author (epark@skku.edu)}
}
\affiliations{
    \textsuperscript{\rm 1}Department of Artificial Intelligence\quad\, \textsuperscript{\rm 2}Department of Mathematics\\ \textsuperscript{\rm 3}Department of Electrical and Computer Engineering\\
    Sungkyunkwan University\\

}

\begin{document}

\maketitle

\begin{abstract}
With the increases in computational power and advances in machine learning, data-driven learning-based methods have gained significant attention in solving PDEs. Physics-informed neural networks (PINNs) have recently emerged and succeeded in various forward and inverse PDE problems thanks to their excellent properties, such as flexibility, mesh-free solutions, and unsupervised training. However, their slower convergence speed and relatively inaccurate solutions often limit their broader applicability in many science and engineering domains. This paper proposes a new kind of data-driven PDEs solver, physics-informed cell representations (PIXEL), elegantly combining classical numerical methods and learning-based approaches. We adopt a grid structure from the numerical methods to improve accuracy and convergence speed and overcome the spectral bias presented in PINNs. Moreover, the proposed method enjoys the same benefits in PINNs, e.g., using the same optimization frameworks to solve both forward and inverse PDE problems and readily enforcing PDE constraints with modern automatic differentiation techniques. We provide experimental results on various challenging PDEs that the original PINNs have struggled with and show that PIXEL achieves fast convergence speed and high accuracy. Project page: https://namgyukang.github.io/PIXEL/
\end{abstract}

\section{Introduction}
Partial differential equations (PDEs) have been central to studying various science and engineering domains~\cite{evans2010partial}. Numerical methods~\cite{smith1985numerical, eymard2000finite} have been developed to approximate the solutions over the centuries. While successful, it requires significant computational resources and domain expertise. As an alternative approach, data-driven machine learning methods have emerged thanks to recent advances in deep learning~\cite{rudy2017data, meng2020ppinn}.

Physics-informed neural networks (PINNs) have recently received significant attention as new data-driven PDE solvers for both forward and inverse problems~\cite{raissi2019physics}. PINNs employ neural networks and gradient-based optimization algorithms to represent and obtain the solutions, leveraging automatic differentiation~\cite{baydin2018automatic} to enforce the physical constraints of underlying PDEs. Although promising and successfully utilized in various forward and inverse problems thanks to its numerous benefits, such as flexibility in handling a wide range of forward and inverse problems and mesh-free solutions, PINNs suffer from slow convergence rates, and they often fall short of the desired accuracy~\cite{krishnapriyan2021characterizing, wang2022respecting, wang2021understanding}.

Training PINNs generally involves deep neural networks and iterative optimization algorithms, e.g., L-BFGS~\cite{liu1989limited} or Adam~\cite{kingma2014adam}, which typically requires a large number of iterations to converge. While many techniques have been developed over the past decades to improve the training efficiency of deep neural networks in general~\cite{girshick2015fast, ioffe2015batch}, high computational complexity is still a primary concern for their broader applicability. 

In addition, multi-layer perceptron (MLP) architecture in low dimensional input domains, where PINNs generally operate, is known to have spectral bias, which prioritizes learning low-frequency components of the target function. Recent studies have shown that spectral bias~\cite{rahaman2019spectral} indeed exists in PINN models~\cite{fbpinns, wang2021eigenvector} and this tendency towards smooth function approximation often leads to failure to accurately capture high-frequency components or singular behaviors in solution functions.

In this paper, we propose physics-informed cell representations (coined as PIXEL), a grid representation that is jointly trained with neural networks to improve convergence rates and accuracy. Inspired by classical numerical solvers that use grid points, we divide solution space into many subspaces and allocate trainable parameters for each cell (or grid point). Each cell is a representation that is further processed by following small neural networks to approximate solution functions. The key motivation behind the proposed method is to disentangle the trainable parameters with respect to the input coordinates. In neural network-only approaches, such as PINNs, all network parameters are affected by the entire input domain space. Therefore, parameter updates for specific input coordinates influence the outputs of other input subspaces. On the other hand, each input coordinate has dedicated trainable parameters updated only for certain input coordinates in \textit{PIXEL}. This parameter separation technique has been explored in visual computing domains~\cite{kanazawaplenoxels, martel2021acorn, reiser2021kilonerf, sun2021direct, muller2022instant,  https://doi.org/10.48550/arxiv.2203.09517} and has shown remarkable success in terms of convergence speed of the training procedure.

Furthermore, the suggested \textit{PIXEL} is immune to spectral bias presented in PINNs. A root cause of the bias is the shared parameters of neural networks for the entire input space. In order to satisfy PDE constraints in all input coordinates, neural networks in PINNs primarily find global principle components of solution functions, usually low-frequency modes. In contrast, PIXEL, each cell is only responsible for a small sub-region of the input domain. Therefore, a large difference between neighboring cell values can better approximate high-frequency components or singular behaviors in PDEs.

Even though we introduce discretized grid representations given the fixed resolution similar to classical numerical methods, such as FEM~\cite{zienkiewicz2005finite}, our approach still enjoys the benefits of PINNs. For example, we can use the same optimization frameworks to solve both forward and inverse PDE problems. Furthermore, PIXEL uses an interpolation scheme to implement virtually infinite resolution grids, and the resulting representations are differentiable with respect to input coordinates. It allows us to enforce PDE constraints using recent \textit{autograd} software libraries~\cite{maclaurin2015autograd, paszke2017automatic}. As a result, our proposed method can be easily plugged into the existing PINNs training pipeline.

In sum, we introduce a new type of PDE solver by combining the best of both worlds, classical numerical methods and automatic differentiation-based neural networks. We use grid representations to improve convergence speed and overcome spectral bias in PINNs. A differentiable interpolation scheme allows us to exploit recent advances in automatic differentiation to enforce PDE constraints. We have tested the proposed method on various PDEs. Experimental results have shown that our approach achieved faster convergence rates and better accuracy.
\section{Related Work}

\textbf{Physics-informed neural network.}
\label{sec:related work}
PINN~\cite{raissi2019physics} is a representative approach that employs a neural network to solve PDEs and operates with few or even without data~\cite{yuan2022pinn}. The main characteristic of PINN is to learn to minimize the PDE residual loss by enforcing physical constraints. In order to compute PDE loss, output fields are automatically differentiated with respect to input coordinates. PINNs are now applied to various disciplines including material science~\cite{lu2020extraction}, and biophysics~\cite{fathi2020super}. Although its wide range of applicability is very promising, it shows slower convergence rates and is vulnerable to the highly nonlinear PDEs~\cite{krishnapriyan2021characterizing, wang2022respecting}.

\textbf{Grid representations.}
This combination of neural networks and grid representations has been explored in other domains. This structure can achieve competitive performance with shallower MLP compared to sole MLP architecture. Since shallower MLP can grant shorter training time to the architecture, novel view synthesis\cite{kanazawaplenoxels, sun2021direct, https://doi.org/10.48550/arxiv.2203.09517}, shape representation\cite{park2019deepsdf, chibane2020implicit}, and image representation\cite{sitzmann2020implicit, chen2021learning, muller2022instant} in computer vision literature enjoy the advantage from the such combined structure. To the best of our knowledge, PIXEL is the first attempt to simultaneously learn grid representations and MLPs to solve challenging linear and nonlinear PDEs.

\textbf{Operator learning.}
Learning mappings from input functions to solution functions has recently gained significant attention in PDE domains.
With the huge success of convolutional neural networks, finite-dimensional operator learning~\cite{guo2016convolutional, zhu2018bayesian} using convolution layers has been studied. To overcome their limitation, e.g., fixed-resolution, \cite{lu2019deeponet, li2020neural, li2020fourier} have proposed to obtain PDE solutions in a resolution invariant manner. Physics-informed operator learning has also been suggested to enforce PDE constraints and further improve the accuracy of solutions~\cite{wang2021learning, li2021physics}. While promising, it requires training datasets primarily obtained from expensive numerical solvers and often suffers from poor out-of-distribution generalization ~\cite{li2020neural}.

\section{PIXEL}

\begin{figure*}[t!]
\begin{center}
\includegraphics[width=0.935\textwidth]{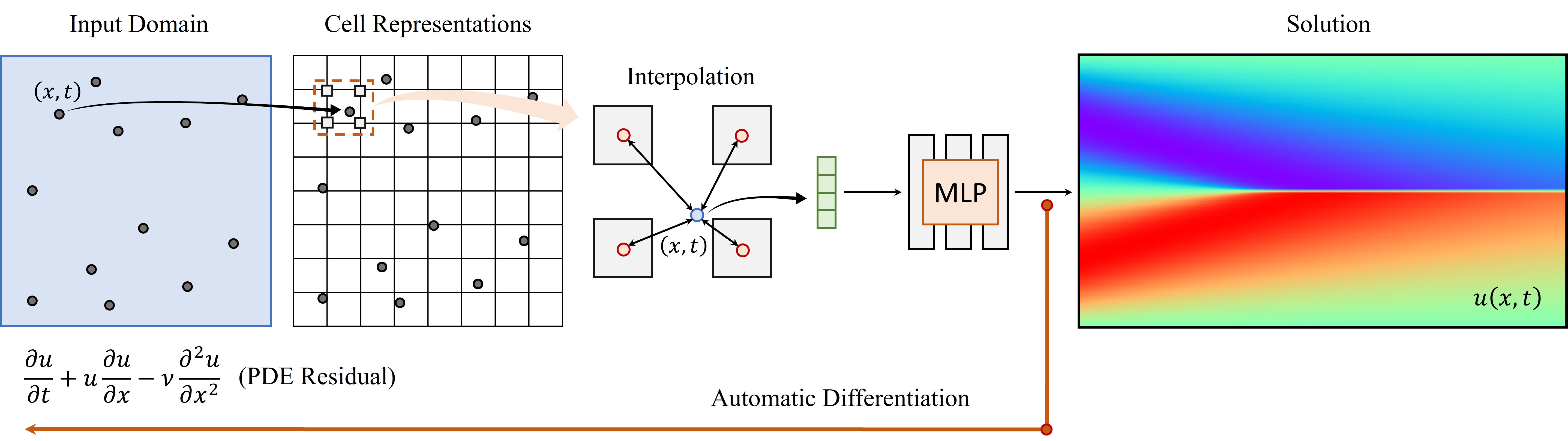}
\end{center}
\caption{Overall PIXEL architecture for a PDE solver}
\label{fig:pixel_architecture}
\end{figure*}


\subsection{Physics-informed neural networks}
We briefly review physics-informed neural networks (PINNs). Let us begin with the initial-boundary value problems with PDEs. A general formulation can be written as follows:
\begin{align}
    & \mathcal{N}_{x,t}[u](x,t) = f(x,t) , \qquad x \in \Omega, t \in [0, T], \\
    & u(x, 0) = g(x),  \qquad \qquad \quad \: x \in \Omega,\\
    & \mathcal{B}_{x,t}[u](x,t) = h(x,t), \qquad \; x \in \partial\Omega, t \in [0, T],
\end{align}
where $\mathcal{N}_{x,t}[\cdot]$ is a linear or non-linear differential operator, $\mathcal{B}_{x,t}[\cdot]$ is also a differential operator for boundary conditions, and the initial conditions are denoted as $u(x, 0) = g(x)$. $u(x,t)$ represents the unknown solution function and PINNs use neural networks, $u_\theta(x,t)$, parameterized by the trainable model parameters $\theta$, to approximate the solution. Then, neural networks are trained by minimizing the following loss function.
\begin{equation}
    \mathcal{L(\theta)} = \lambda_\text{res}\mathcal{L}_\text{res}(\theta) + \lambda_\text{ic}\mathcal{L}_\text{ic}(\theta) + \lambda_\text{bc}\mathcal{L}_\text{bc}(\theta) + \lambda_\text{data}\mathcal{L}_\text{data}(\theta),
\end{equation}
where, $\mathcal{L}_\text{res}, \mathcal{L}_\text{ic}, \mathcal{L}_\text{bc}, \mathcal{L}_\text{data}$ are PDE residual, initial condition, boundary condition, and observational data loss functions, respectively. $\lambda$ are weighting factors for each loss term. Each loss term is usually defined as mean square loss functions over sampled points. For example, a PDE residual loss $\mathcal{L}_\text{res}$ over $N_\text{res}$ collocation points can be written as,
\begin{equation}
    \mathcal{L}_\text{res}(\theta) = \frac{1}{N_\text{res}}\sum_{i=1}^{N_\text{res}} |\mathcal{N}_{x,t}[u_\theta](x_i, t_i) - f(x_i, t_i)|^2.
\end{equation}
For forward problems, observational data is not generally available, hence $\lambda_\text{data}=0$. In contrast, observational data is accessible in inverse problems, and initial and boundary conditions may also be available depending on the cases~\cite{raissi2018deep, raissi2019physics}. Gradient-based optimization algorithms are used to optimize the loss function, and L-BFGS and Adam are widely used in PINNs literature. Automatic differentiation is used to compute the gradients of both differential operators w.r.t input coordinates, and the loss function w.r.t trainable neural network parameters.

\subsection{Neural networks and grid representations}
The proposed architecture consists of a small neural network and a feature extractor of input coordinates using grid representations. We approximate solution functions by a neural network $f$ parameterized by $\theta$,
\begin{equation}
    u(x,t) \approx f(\phi(\hat{x},\hat{t},\mathcal{C});\theta),
\end{equation}
where $\mathcal{C}$ is a grid representation and $\phi$ is a feature extractor given input coordinates and the grid representation using an interpolation scheme, which will be further explained in the next section. Note that both $\mathcal{C}$ and $\theta$ are model parameters and are updated during the training procedure. The dimension of $\mathcal{C}$ is determined by the dimension of the spatial input domain. For example, if $x \in \Omega \subset \mathbb{R}$ then $\mathcal{C} \in \mathbb{R}^{c \times H \times W}$ is a three dimensional tensor, where the channel size $c$, and $H$ and $W$ are spatial and temporal grid sizes, respectively. $\hat{x} \in [1,H]$ and $\hat{t}\in [1,W]$ are normalized input coordinates assuming input domain $\Omega \subset \mathbb{R}$ and $[0,T]$ are tightly bounded by a rectangular grid.
If $x \in \Omega \subset \mathbb{R}^2$ then $\mathcal{C}$ is a four dimensional tensor, and if $x \in \Omega \subset \mathbb{R}^3$ then $\mathcal{C}$ is a five dimensional tensor~\footnote{Without temporal coordinates, e.g., Helmholtz equation, $\mathcal{C}$ is three or four dimensional tensors, respectively.}.

The proposed grid representation is similar in spirit to classical numerical methods such as FDM~\cite{smith1985numerical} or FEM~\cite{zienkiewicz2005finite}, which can increase the accuracy of solutions by extending the grid size or using more fine-grained nodal points. PIXEL inherits this advantage, and we can obtain the desired accuracy and better capture high-frequency details of solution functions by larger grid representations. Furthermore, we learn representations at each nodal point instead of directly approximating solution fields. A neural network further processes them to obtain the final solutions, which enables us to express a more complex and richer family of solution functions.

\subsection{Mesh-agnostic representations through interpolation}
In two dimensional grid cases, $x \in \Omega \subset \mathbb{R}$ and $\mathcal{C} \in \mathbb{R}^{c \times H \times W}$, the following is a feature extractor from the cell representations,
\begin{align}
    \phi(\hat{x},\hat{t},\mathcal{C})& = \\\sum_{i=1}^{H} \sum_{j=1}^{W} &\mathcal{C}_{ij}k(\text{max}(0,1-|\hat{x}-i|))k(\text{max}(0,1-|\hat{t}-j|)),\nonumber
\end{align}
where $\mathcal{C}_{ij} \in \mathbb{R}^{c}$ denotes cell representations at ${(i,j)}$, and $k: [0,1] \rightarrow [0,1]$ represents a monotonically increasing smooth function. Given normalized coordinates $(\hat{x},\hat{t})$, it looks up neighboring points ($2^{d+1}$ points) and computes the weighted sum of the representations according to a predefined kernel function. It is differentiable w.r.t input coordinates so that we can easily compute partial derivatives for PDE differential operator $\mathcal{N}[\cdot]$ by using automatic differentiation. In the context of neural networks, it was used in a differentiable image sampling layer~\cite{jaderberg2015}, and this technique has been extensively explored in various domains~\cite{dai2017deformable, he2017mask, pedersoli2017areas}. Although not presented, this formulation can be easily extended to higher dimensional cases.

To support higher-order gradients, we need to use a kernel function multiple times differentiable depending on governing PDEs. We use a cosine interpolation kernel, $k(x) := \frac{1}{2}(1-\cos(\pi x))$ because it is everywhere continuous and infinitely differentiable. We empirically found that it is superior to other choices, e.g., RBF kernel, in terms of computational complexity and solution accuracy. A bilinear interpolation by $k(x):=x$ is still a valid option if PDE only contains the first order derivatives and the goal is to maximize computational efficiency.

PINNs have been widely adopted in various PDEs due to their mesh-free solutions, which enable us to deal with arbitrary input domains. Even though we introduce grid representations, our proposed method is also not limited to rectangular input domains. We use a differentiable interpolation scheme, and we can extract cell representations at any query point. In other words, we connect the discretized grid representations in a differentiable manner, and the grid resolution becomes infinity. The predetermined cell locations might affect the solution accuracy meaningfully~\cite{mekchay2005convergence, prato2019adaptive}, and a more flexible mesh construction would be ideal. We believe this is an exciting research direction, leaving it as future work.

\subsection{Multigrid representations}
\label{multigrid representations}
Since we introduce the grid representation that each grid point covers only small subregions of the input domain, the higher the resolution of the grid, the faster convergence and more accurate solutions are expected. However, a significant challenge of scaling up the method is that the number of training data points required will likely increase exponentially. Given randomly sampled collocation points, the more fine-grained grids, the less chance a grid cell would see the points during the training. It would result in highly overfitted solutions since finding a satisfactory solution in a small local region with only a few data points would easily fall into local minima e.g., PDE residual loss is very low, but generating wrong solutions. Furthermore, since we no longer rely on the neural network's smooth prior, the proposed grid representations suffer from another spectral bias, which tends to learn high-frequency components quickly. Due to these reasons, the solution function we obtained from high-resolution grids often looks like pepper noise images while quickly minimizing the PDE loss.

\begin{figure}[t!]
\begin{center}
  \includegraphics[width=0.45\textwidth]{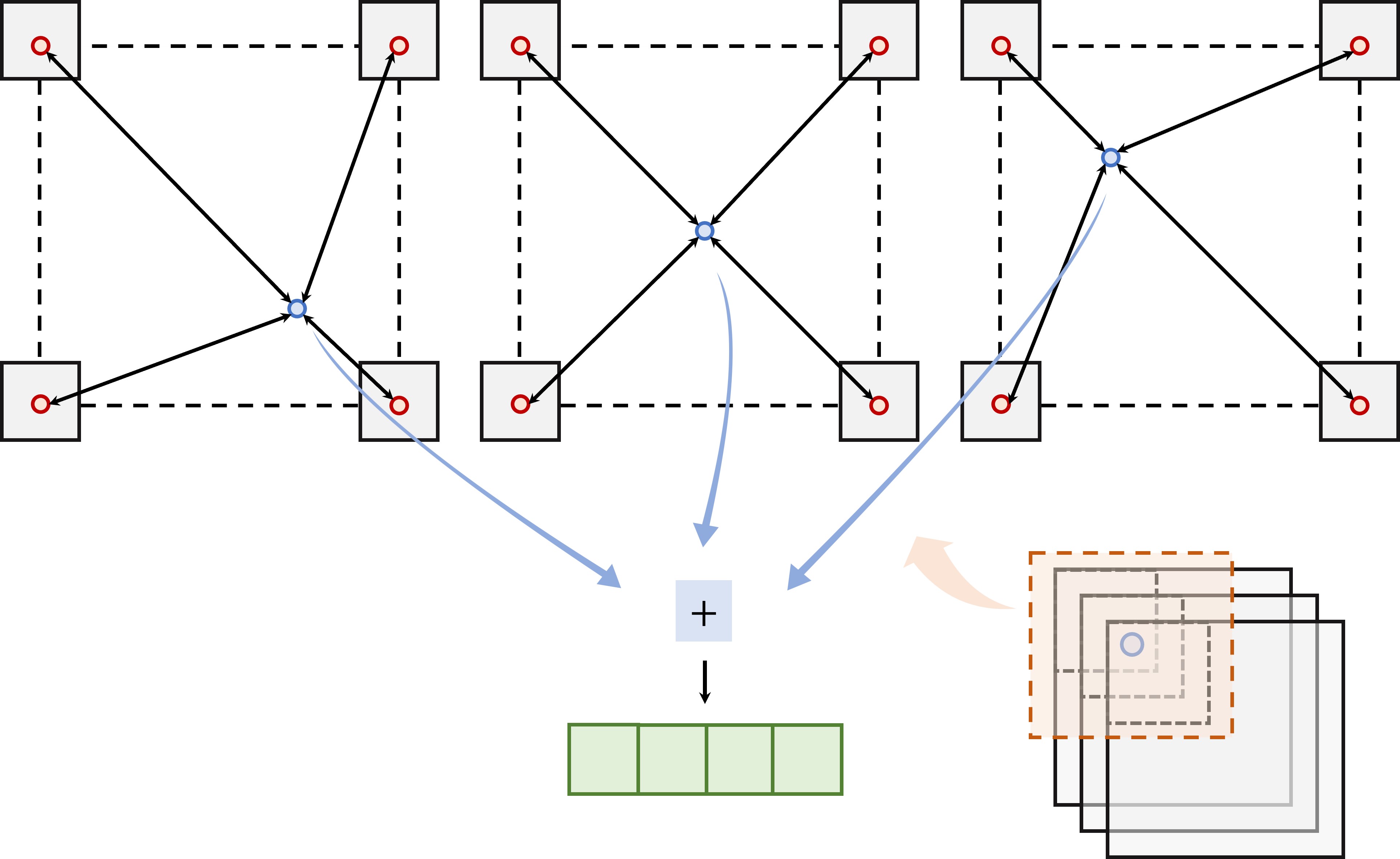}
  \end{center}
  \caption{Multigrid representations}
  \label{fig:multicell}
\end{figure}

One way to inject a smooth prior and avoid overfitting is to look up more neighboring cells for the interpolation, such as cubic interpolation or other variants, instead of the suggested scheme looking at only corner cells in squares or hypercubes. By doing so, more neighboring grid cells are used to compute representations at one collocation point, which would reduce the overfitting issues discussed above. However, it is computationally expensive, and the number of neighboring cells required to look up will vastly increase according to the grid sizes and dimensions.

Inspired by recent hierarchical grid representations~\cite{takikawa2021nglod,muller2022instant}, we suggest using multigrid representations. We stack up multiple coarse-grained grid representations and the representations at each collocation point are computed by summing up the representations from all grids. With a slight abuse of notation, a multigrid representation is defined as four dimensional tensors in two dimensional grids $\mathcal{C} \in \mathbb{R}^{M \times c \times H \times W}$. Then we can reformulate an interpolation function as,
\begin{equation}
    \phi_{\text{multi}}(\hat{x},\hat{t},\mathcal{C}) = \sum_{i=1}^{M} \phi(\hat{x}+\frac{(i-1)}{M},\hat{t}+\frac{(i-1)}{M}, \mathcal{C}^i),
\end{equation}
where $C^i \in \mathbb{R}^{c \times H \times W}$ denotes a grid representation in equation 8. We present a pictorial demonstration of multigrid representations in Figure~\ref{fig:multicell}. We have $M$ grids, and each grid is shifted in such a way that an input coordinate can be located in different locations in each grid. In this way, we can increase the effective grid size by a factor of $M$, meaning the model's expressive power will also increase. Without shifting each grid, an input coordinate lies on identical locations in every grid, and each grid would represent similar values resulting in not increasing the expressive power. The suggested multigrid approach was very critical to overall PIXEL performance. Without this, we observed that PIXEL suffers from serious overfitting issues.

\section{Experiments}
This section compares PIXEL with the baseline model PINN on the various linear and nonlinear PDEs. First, we provide a motivating example of what a baseline PINN suffers from, where the solution functions contain high-frequency components. Then, we experimented on various linear PDEs, such as 1D Convection, Reaction-diffusion, and Helmholtz equation. We also tested PIXEL on the Allen-Cahn and Burgers equation to test our model's capability to solve non-linear PDEs. For all experiments, we used Limited-memory BFGS (L-BFGS) second-order optimization algorithms. To compute the accuracy of the approximated solutions, we used the relative $L_2$ error, defined as $\frac{||u-\hat{u}||_2}{||u||_2}$, where $\hat{u}$ is a predicted solution and $u$ is a reference solution. \textit{Experimental details will be provided in supplementary materials.}


\begin{table*}[ht]
\centering
\def\arraystretch{1.1}%
\small{
\begin{tabular}{l|c|c|c|c}
\toprule
& \multirow{2}{*}{{PDEs}} & \multirow{2}{*}{{Initial condition}} & \multirow{2}{*}{{Boundary condition}} & {Inverse problem}\\ 
& & & &  {coefficient} \\\hline 
\multirow{2}{*}{Convection}  & $u_t + \beta u_x = 0$ & \multirow{2}{*}{$u(x, 0) = \sin x$} & \multirow{2}{*}{$u(0,t) = u(2\pi, t)$} & \multirow{2}{*}{$\beta$} \\
& $x \in [0, 2\pi], ~t \in [0, T]$ & & \\ \hline 
Reaction & $u_t - \nu u_{xx} - \rho u(1-u)=0$ & \multirow{2}{*}{$u(x,0) = h(x)$} &\multirow{2}{*}{$u(0,t) = u(2\pi, t)$}  & \multirow{2}{*}{$\nu$}\\
Diffusion &  $ x \in [0, 2\pi], t \in [0,T]$ & & \\ \hline 
\multirow{5}{*}{Helmholtz}  & $\Delta u(x, y, z) + k^2u(x, y, z) = q(x,y,z)$ & & \multirow{3}{*}{$u(x,y,z) = 0,$}& \\
& $q(x,y,z)= k^2 \sin{(a_1 \pi x)}\sin{(a_2 \pi y)}\sin{(a_2 \pi z)} $ & & \\
  \multirow{3}{*}{\quad(3D)}& $- (a_1 \pi)^2  \sin{(a_1 \pi x)}\sin{(a_2 \pi y)}\sin{(a_2 \pi z)}$&$\cdot$ & \multirow{3}{*}{$(x,y,z) \in \partial[-1, 1]^2$}   &$\cdot$  \\
& $~ -(a_2 \pi)^2 \sin{(a_1 \pi x)}\sin{(a_2 \pi y)}\sin{(a_2 \pi z)} $  & &\\ 
&  $(x,y,z) \in [-1, 1]^2$ & & \\ \hline
\;\, Navier & $u_t + \lambda_1 (uu_x +vu_y) = -p_x + \lambda_2(u_xx + u_yy)$ &  \multirow{3}{*}{$\cdot$} &  \multirow{3}{*}{$\cdot$} &  \multirow{3}{*}{$\lambda_1, \lambda_2$}\\
\;\,  -Stokes & $v_t + \lambda_1 (uv_x +vv_y) = -p_y + \lambda_2(v_xx + v_yy)$& &  &\\ 
\, \quad(3D)& $u_x + v_y = 0$ & &  &\\ \hline 
\multirow{2}{*}{Allen-Cahn} & $u_t - 0.0001u_{xx} + \lambda u^3 - 5u = 0$ & \multirow{2}{*}{$u(x,0) = x^2 \cos(\pi x)$} & $u(t,-1) = u(t,1)$ &  \multirow{2}{*}{$\lambda$}\\
& $ x \in [-1,1], t \in [0,1], \lambda = 5  $ & & $u_x(t,-1) = u_x(t,1)$ &\\ \hline 
\multirow{2}{*}{Burgers} & $u_t + uu_x - \nu u_{xx} = 0$ & \multirow{2}{*}{$u(0,x) = -\sin (\pi x)$} & \multirow{2}{*}{$u(t,-1) = u(t,1) = 0$} & \multirow{2}{*}{$\nu$} \\
& $x \in [-1,1], ~t \in [0,1]$ & & \\
\bottomrule
\end{tabular}
}
\caption {The formulations of various PDEs in our experiments of the forward and the inverse problem.}
\label{table:pdes}
\end{table*}

\subsection{Implementation}
We implemented the 2D, and 3D customized CUDA kernel of the triple backward grid sampler that supports cosine, linear, and smoothstep kernel~\cite{muller2022instant} and third-order gradients $u_{xxc}, u_{yyc}$ with second-order gradients~\cite{wang2022go}. As a result, the runtime and the memory requirement were significantly reduced. You can find our customized CUDA kernel code at https://github.com/NamGyuKang/CosineSampler.
\subsection{An illustrative example}
\begin{figure}[hb!]
  \includegraphics[width=0.48\textwidth]{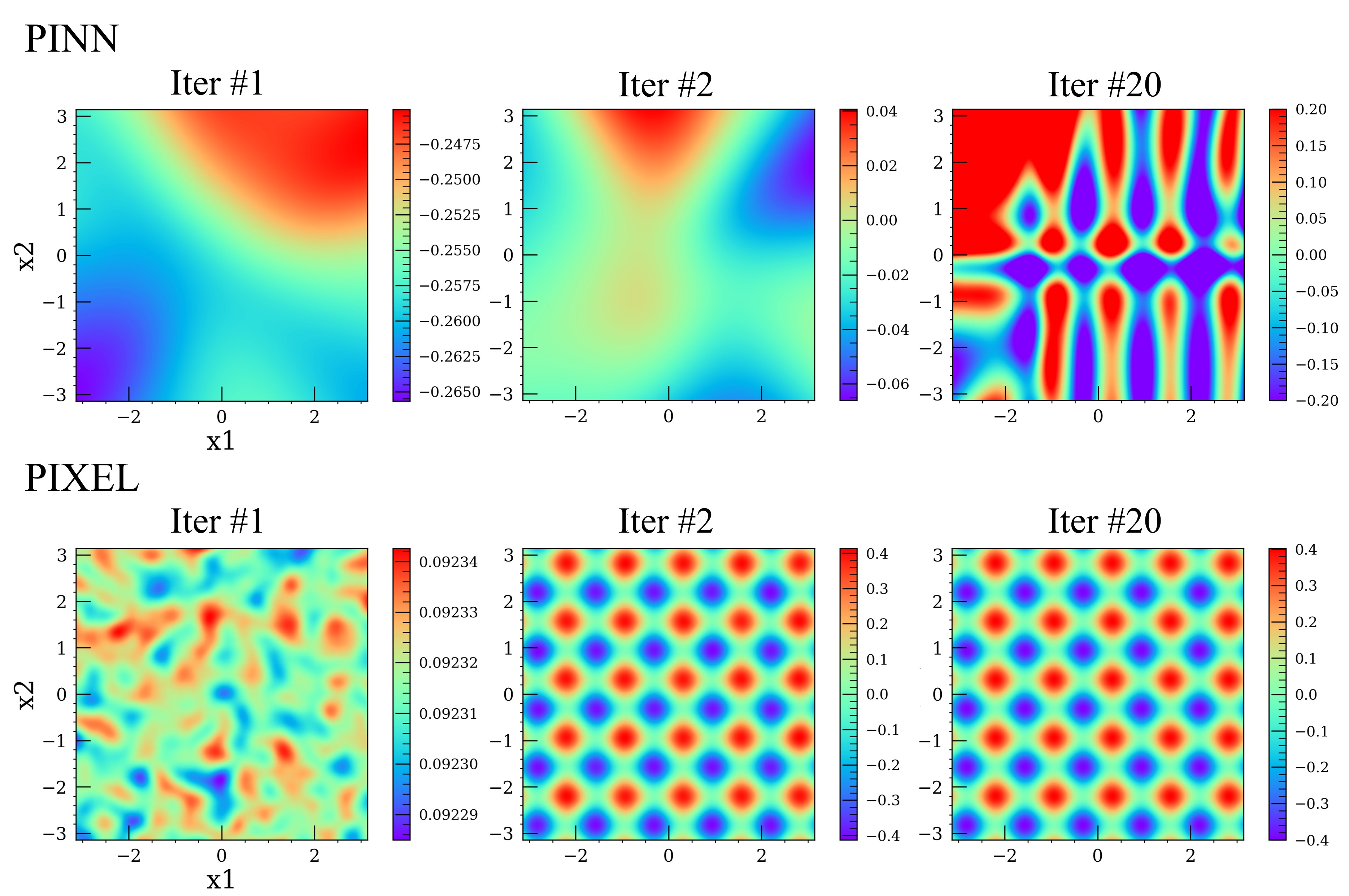}
  \caption{An illustrative sinusoid example}
  \label{fig:sinusoid}
\end{figure}

We begin with a motivating example that PINNs have struggled to find an accurate solution, suggested in ~\cite{fbpinns}. It is the first-order linear PDE and has high-frequency signals in the exact solution, $\frac{\partial u}{\partial x_1}+\frac{\partial u}{\partial x_2} = \cos(\omega x_1) + \cos(\omega x_2), (x_1,x_2) \in [2\pi, 2\pi]^2, u \in \mathbb{R}$. $\omega$ controls frequency in solutions, and to test the capability of PIXEL to capture complex high-frequency solutions, we set $\omega=15$. As the results show, our method converges very rapidly. Indeed, we could find an accurate solution in a few iterations, and we already see a clear solution image in \textbf{\textit{two}} L-BFGS iteration (Figure~\ref{fig:sinusoid}). On the other hand, PINN could not find a satisfactory solution after many iterations even though we have tested numerous hyperparameters. We believe it validates that the proposed method converges very fast and does not have a spectral bias that neural networks commonly encounter.

\subsection{PDEs description}
\label{sec:experimental_setup}
\quad\textbf{1D convection equation.}
The convection equation describes the heat transfer attributed to the fluid movement. \cite{krishnapriyan2021characterizing} has studied this PDE in the context of PINNs and without sequential training, the original PINNs has failed to find accurate solutions. We used $\beta=30$ and the initial and boundary conditions of all PDEs used for the experiments are described in Table~\ref{table:pdes}. 

\textbf{Reaction-diffusion equation.}
Reaction-diffusion equation is also a PDE that the original PINNs have worked poorly~\cite{krishnapriyan2021characterizing}. We used the same formulation in \cite{krishnapriyan2021characterizing}, and conducted experiments with the same PDE parameters ($\rho=5$, $\nu=3$). 

\textbf{Helmholtz equation.}
Helmholtz equation describes the problems in the field of natural and engineering sciences like acoustic and elastic wave propagation. We used the same formulation in \cite{wang2021understanding}. 
~\cite{wang2021understanding} has reported that the original PINN has struggled to find an accurate solution and proposed a learning rate annealing algorithm and a novel neural network architecture. 

\textbf{Allen-Cahn equation.}
Allen-Cahn equation is a non-linear second-order PDE that is known to be challenging to solve using conventional PINNs~\cite{wang2021understanding}, and a few techniques, including adaptive re-sampling~\cite{adaptivepinn} and weighting algorithms~\cite{softattention, liu2021dual, wang2022and}, have been proposed. 

\textbf{1D Burgers equation.}
Finally, we also conducted experiments on 1D Burgers equation. It is a standard benchmark in PINNs literature, which is known to have a singular behavior in the solution. We used the same PDE parameter, $\nu = 0.01/\pi$ in \cite{raissi2019physics}. 

\textbf{3D Navier-Stokes eqation}
 The non-linear second order 3D Navier-Stokes is well known as a challenging equation to solve for fluid dynamics.
 \cite{raissi2019physics} shows the result of the inverse problem, which is the multivariate coefficient simultaneous prediction. We predicted the same coefficients, $\lambda_1 = 1.0, \lambda_2 = 0.01$.

\begin{figure*} [ht]
    \centering
    \includegraphics[width=\textwidth]{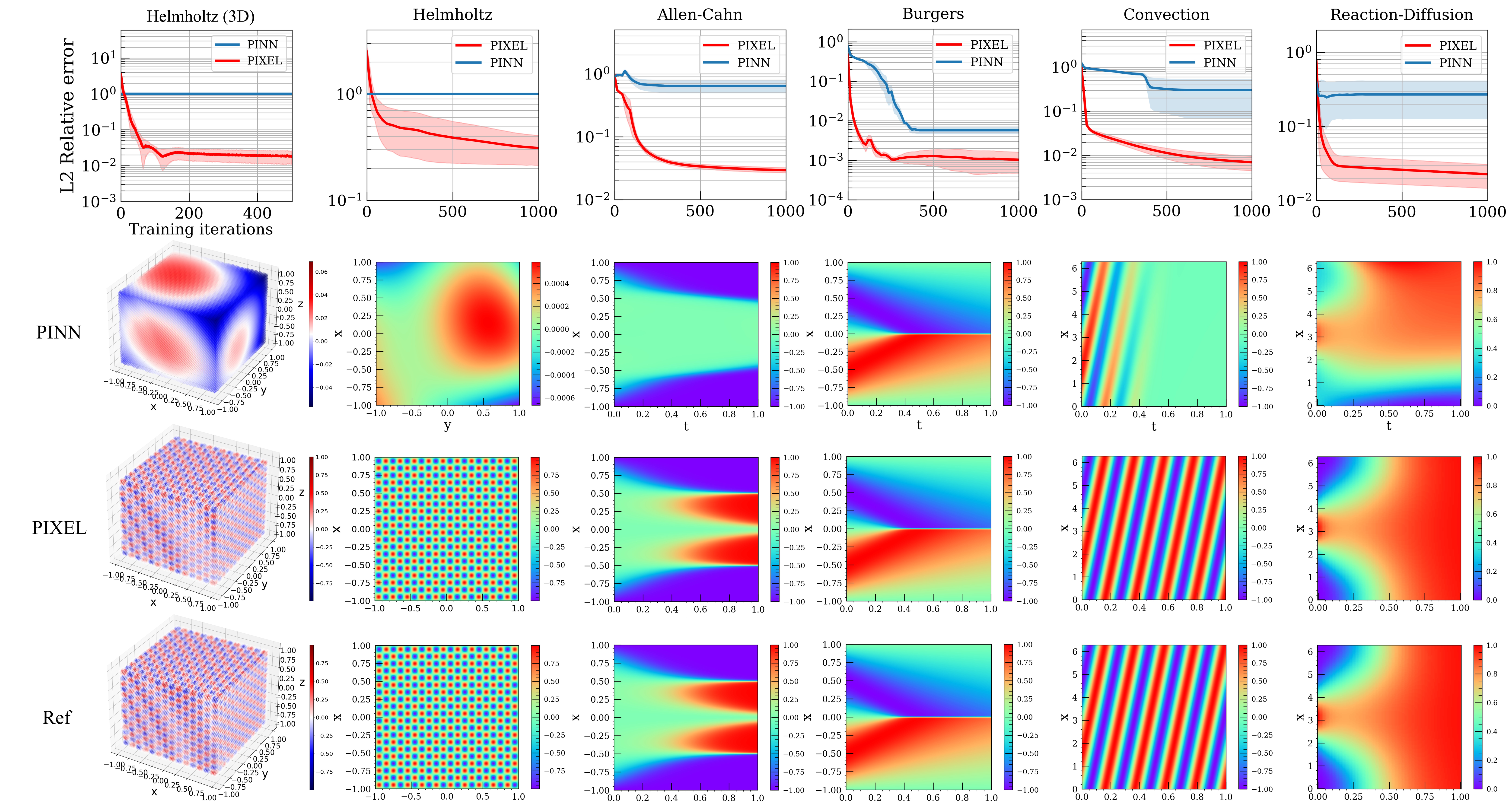}
    \caption{For the forward problem, training loss curves and solutions of various PDEs: We run both PINN and PIXEL 5 times for each PDE experiment, and the shaded areas show $80\%$ confidence interval of 5 different runs with different \textit{random} seeds (100, 200, 300, 400, 500). Each PDE parameters are followed. $\beta=30$ in convection, $\nu=3, \rho=5$ in reaction diffusion, $a_1=7, a_2=7, a_3=7, k=1$ in Helmholtz (3D), $a_1=10, a_2=10, k=1$ in Helmholtz (2D), and $\nu = 0.01/\pi$ in Burgers equations. We showed solution images of the best performing one out of 5 different runs.}
    \label{fig:loss_curves}
\end{figure*}

\begin{figure*}[t!] 
    \centering
    \includegraphics[width=\textwidth]{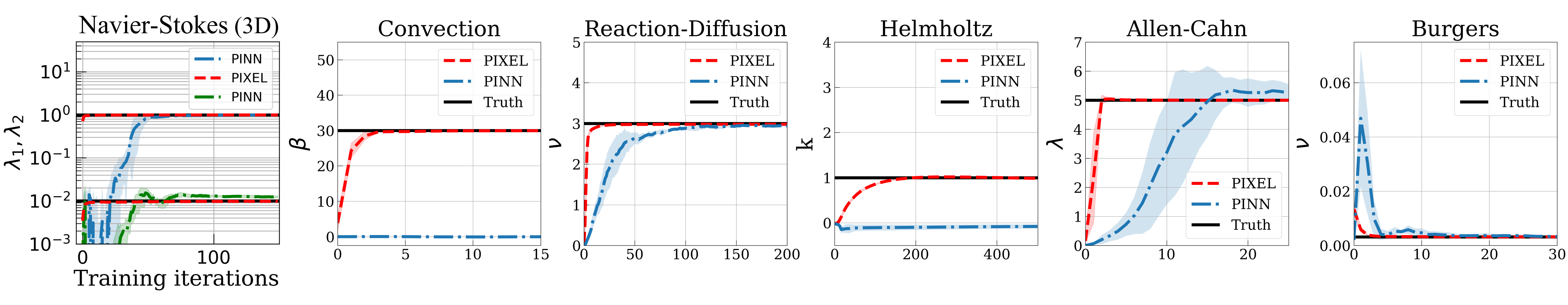} 
    \caption{Experimental results of the inverse problems. PIXEL shows a faster convergence speed compared to PINN and more accurate PDE parameter predictions. The shaded areas of training curves show $95\%$ confidence interval of 5 different runs with different \textit{random} seeds (100, 200, 300, 400, 500).}
    \label{fig:inverse_seed}
\end{figure*}

\begin{table*}[t!]
\centering
\small{
\begin{tabular}{l|c|c|c|c|c|c}
\toprule
Methods & Helmholtz (3D) & Helmholtz (2D) & Allen-Cahn & Burgers & Convection & Reaction \\
& ($a_1,a_2,a_3=7$)&  ($a_1,a_2=10$)  & & &  & diffusion \\ \hline
PINN (8-40) & N/A & N/A & N/A & 5.60e-04 & N/A & N/A \\ \hline 
Sequential & \multirow{2}{*}{N/A} &\multirow{2}{*}{N/A}  &\multirow{2}{*}{N/A}  &\multirow{2}{*}{N/A}  &\multirow{2}{*}{2.02-e02}&\multirow{2}{*}{1.56e-02}\\ 
training & & & & & & \\ \hline 
Self-attention  & N/A & N/A & 2.10e-02 & N/A& N/A & N/A \\ \hline 
Time marching & N/A & N/A & 1.68e-02 & N/A& N/A & N/A \\ \hline 
Causal training & N/A & N/A & 1.43e-03 & N/A& N/A & N/A \\ \hline
Causal training  & \multirow{2}{*}{N/A}  &\multirow{2}{*}{N/A}& \multirow{2}{*}{\textbf{1.39e-04}} &\multirow{2}{*}{N/A}&\multirow{2}{*}{N/A}&\multirow{2}{*}{N/A} \\ 
(modified MLP)& & & & & & \\ \hline
 &  1.00 & 1.00 & 9.08e-01 & 5.77e-03 &3.02e-01 &  2.46e-01  \\
PINN (ours) & ($\pm$ 7.19e-04) & ($\pm$ 1.49e-06) & ($\pm$ 1.68e-02) & ($\pm$ 1.74e-03) & ($\pm$ 3.40e-01) & ($\pm$ 2.25e-01) \\
 & (best: 1.00) &(best : 1.00) & (best : 5.23e-01) & (best : 3.35e-03) & (best : 2.45e-02) & (best : 2.36e-02) \\ 
\hline 
 \multirow{2}{*}{PIXEL} & 5.06e-03 & 3.05e-01 & 1.77e-02 &  9.98e-04 & 9.48e-03  & 1.63e-02 \\
\multirow{2}{*}{(16,4,16,16)} & ($\pm$ 2.64e-03)&  ($\pm$ 2.38e-01) &  ($\pm$ 4.67e-03) &($\pm$ 3.70e-04)&($\pm$ 1.62e-03)& ($\pm$ 2.11e-03) \\ 
& \textbf{(best : 6.61e-04)} & \textbf{(best : 7.47e-02)} & (best : 9.64e-03) & (best : 4.88e-04) & (best : 6.39e-03) & (best : 1.33e-02)\\ \hline 
  \multirow{2}{*}{PIXEL}  & 1.95e-01 & 4.26e-01 & 1.90e-02 & 6.20e-04 & 4.69e-03  & 8.11e-03  \\ 
\multirow{2}{*}{(64,4,16,16)}& ($\pm$ 1.84e-01) &($\pm$ 3.10e-01) & ($\pm$ 8.35e-03) & ($\pm$ 2.09e-04) & ($\pm$ 1.25e-03)& ($\pm$ 8.74e-05) \\
& (best : 4.86e-02)& (best : 1.05e-01) & (best : 3.85e-03) & (best : 3.85e-04)& (best : 2.41e-03) & (best : 7.81e-03) \\\hline 
  \multirow{2}{*}{PIXEL}  & 1.53e-01 & 3.11e-01 & 1.63e-02 & 7.01e-04& 6.19e-03 &  8.26e-03\\ 
\multirow{2}{*}{(96,4,16,16)} &($\pm$ 6.81e-02) &($\pm$ 1.43e-01)  & ($\pm$ 3.95e-03) & ($\pm$ 3.60e-04)& ($\pm$ 3.36e-03) &($\pm$ 1.18e-03) \\ 
& (best : 1.34e-02)& (best : 1.70e-01) & (best : 8.86e-03) & \textbf{(best : 3.20e-04)}& \textbf{(best : 1.84e-03)} & \textbf{(best : 7.12e-03)}  \\
\bottomrule
\end{tabular}}
\caption {The comparisons to other methods ($L_2$ relative errors). Five different experiments were performed and averaged. The standard deviation is shown with the mean in the table. Seeds 100, 200, 300, 400, and 500 were used. we compared against PINN~\cite{raissi2019physics}, Sequential training~\cite{krishnapriyan2021characterizing}, Self-attention~\cite{softattention}, Time marching~\cite{mattey2022novel}, and Causal training~\cite{wang2022respecting}.}
\label{table:l2_comparison}
\end{table*}

\subsection{Results and discussion}
As in Figure~\ref{fig:loss_curves}, our method converges faster than the PINN baseline in the number of L-BFGS training iterations. In all cases, our method obtained accurate solutions (indistinguishable from the reference solutions) in a few tens of iterations. We showed the training loss curves over 1000 iterations. 

For the forward problem, In Convection equation, we observed the same phenomena in \cite{krishnapriyan2021characterizing}. The baseline method PINN converges slowly, and the resulting solution image was not correctly updated for the later time domain, $t>0.4$. For Reaction-diffusion equation, PINN showed a constant curve shape after 285 iterations in the averaged loss curve. Whereas PIXEL showed an exponential decay shape until 10,000 iterations.

In both 2D and 3D Helmholtz, we used high-frequency parameters $a_1=10, a_2=10$ for 2D, and $a_1=7, a_2=7, a_3=7$ for 3D. which resulted in a very complex solution. As we expected, due to the spectral bias, PINN has failed to converge to an accurate solution. In contrast, PIXEL obtained a highly accurate solution quickly. With low-frequency parameters $a_1=1, a_2=4$ in 2D, PIXEL achieved the best performance result (8.63e-04) with 96 multigrids. However, PINN got a higher relative $L_2$ error (2.30e-03) than PIXEL.

 For Allen-Cahn, which is known to be notoriously difficult, the previous studies have demonstrated that PINNs perform very poorly without additional training techniques, such as time marching techniques ~\cite{mattey2022novel} or causal training~\cite{wang2022respecting}. However, our method can obtain accurate solutions without any additional methods.

For the inverse problem, PINN showed fluctuation in the prediction curve due to the high standard deviation by the \textit{random} seed. On the other hand, PIXEL showed robustness in predicting regardless of \textit{random} seed in 3D as well as 2D equations within a $95\%$ confidence interval. Except for Helmholtz equation which PINN failed to train, PIXEL showed convergence in a \textit{few} iterations for PDEs.

To compare recent advanced PINNs algorithms, we also provide quantitative results in Table~\ref{table:l2_comparison}. We reported the numbers from the original papers and we also provided our implementation of the baseline PINN method denoted as PINN (ours). For higher accuracy, we trained more iterations until convergence, 10k, 1k, 500k, 39k, 18k, and 10k iterations were performed for each PDE according to the sequence shown in the 
Table~\ref{table:l2_comparison}, respectively. Our method outperforms other methods for all PDEs, except Allen-Cahn equation. Recently proposed causal PINN~\cite{wang2022respecting}, they proposed a loss function that gives more weights to the data points that have already reached low loss values. We can also incorporate this technique into our methods. However, this paper aims to show the performance of the newly proposed architecture without any bells and whistles, and combining recent training techniques into our framework will be a promising research direction.

\section{Conclusion, limitations, and discussion}
\label{sec:limitations}
We propose a new learning-based PDE solver, PIXEL, combining the conventional grid representations and recent neural networks-based solvers. We showed that PIXEL converges rapidly and approximates the solutions accurately on various challenging PDEs. We believe this approach bridges the gap between classical numerical methods and deep learning, especially recently popularized neural fields~\cite{neuralfields}. 

While promising, it would be more convincing if we provided experiments on PDEs with higher-order derivatives such as Kuramoto-Sivashinsky, and Sawada-Kotera equations. Early results showed that both PIXEL and PINNs are converging very slowly, and we hypothesize that using automatic differentiation suffers from the vanishing gradients problem. We plan to further investigate this phenomenon.

Another natural question related to spatial complexity may arise. The proposed grid-based representations would require intolerable memory footprints for higher dimensional PDEs, such as BGK-Boltzmann equations. To achieve high accuracy, we may need to increase the grid size arbitrarily large. We believe these are open and exciting questions, and combining numerical methods and machine learning techniques may come to the rescue. For example, tensor decomposition techniques~\cite{tensordecomp, https://doi.org/10.48550/arxiv.2203.09517}, data compression algorithm~\cite{le1991mpeg, jpeg}, or adaptive methods~\cite{adaptivefem} are few good candidates. We believe we provide a good example that combines neural networks and grid representations for solving PDEs, and marrying existing techniques into the proposed algorithm will be an exciting research area.

\section*{Acknowledgements}
We are thankful to Junwoo Cho for helpful discussion and contributions. This research was supported by the Ministry of Science and ICT (MSIT) of Korea, under the National Research Foundation (NRF) grant (2022R1F1A1064184, 2022R1A4A3033571), Institute of Information and Communication Technology Planning Evaluation (IITP) grants for the AI Graduate School program (IITP-2019-0-00421). The research of Seok-Bae Yun was supported by Samsung Science and Technology Foundation under Project Number SSTF-BA1801-02.
\fontsize{9.8pt}{10.8pt} \selectfont

\bibliography{aaai23}

\clearpage
\appendix

\section{Experimental setup and details}
For all experiments, we used Limited-memory BFGS (L-BFGS) second-order optimization algorithms. In many cases of training PINNs, it outperforms other first-order optimization algorithms, such as ADAM or SGD. We set the learning rate to 1.0 and used the strong-wolfe line search algorithm. In every L-BFGS iteration, we randomly sample collocation points to make the model robust to the entire input and time domains for training PIXELs. We found that PINNs often have struggled to converge in this setting, so we initially sampled the collocation points and fixed them, which has been a common practice in PINNs literature. To compute the accuracy of the approximated solutions, we used the relative $L_2$ error, defined as $\frac{||u-\hat{u}||_2}{||u||_2}$, where $\hat{u}$ is a predicted solution and $u$ is a reference solution. We used NVIDIA RTX3090 GPUs and A100 GPUs with 40 GB of memory. For all experiments, we used 1 hidden layers and 16 hidden dimensions for  shallow MLP architecture, and a hyperbolic tangent activation function (tanh) was used. For coefficients of the loss function, we used $\lambda_{\text{ic}}=\lambda_{\text{bc}}=1, \lambda_{\text{data}}=0$. 

\textbf{1D convection equation.}
A shallow MLP of 2 layers with 16 hidden units was used. The baseline PINN model was trained with the same number of data points from PIXEL. For the PINN model, we used 3 hidden layers and 50 hidden dimensions following the architecture in \cite{krishnapriyan2021characterizing}.

\textbf{Reaction-diffusion equation.}
For training PINNs, we used 3 hidden layers and 50 hidden dimensions following the architecture in \cite{krishnapriyan2021characterizing}.

\textbf{Helmholtz equation.}
the source term is given as $q(x, y) = -(a_1 \pi)^2 u(x, y) - (a_2 \pi)^2 u(x, y) + k^2 u(x, y)$. The analytic solution of this formulation is known as $u(x, y) = \sin{(a_1 \pi x)}\sin{(a_2 \pi y)}$. We tested the PDE parameters $k=1$, $a_1=1$ ,and $a_2=4$. For a more complex setting, we also tested $k=1$, $a_1=10$ and $a_2=10$. For the baseline PINN model, we used 7 hidden layers and 100 hidden dimensions following the architecture in \cite{wang2021understanding}.

\textbf{Allen-Cahn equation.}
For the baseline PINN model, we used 6 hidden layers and 128 hidden dimensions following the architecture in \cite{mattey2022novel} In the case of Allen-Cahn, there was a problem that NaN occurs in PINN when the seed is 400. Unlike PIXEL, PINN excludes seed 400 only in the case of Allen-Cahn.

\textbf{1D Burgers equation.}
For the baseline PINN model, we adopted the same architecture in \cite{raissi2019physics}, using 8 hidden layers and 40 hidden dimensions.

\subsection{Hyperparameter of experiments}
\begin{table*}[t!]
\centering
\begin{tabular}{l|c|c|c|c|c}
\toprule
& Convection & Reaction & Helmholtz & Allen-Cahn & Burgers \\ & & diffusion & & & \\ \hline
Grid sizes &  (96, 4, 16, 16) &  (96, 4, 16, 16) &  (96, 4, 16, 16)& (96, 4, 16, 16) & (96, 4, 16, 16) \\ \hline 
\# collocation pts & 100,000 & 100,000 & 100,000 & 100,000 & 100,000 \\ \hline
\# ic pts & 100,000 & 100,000 & N/A & 100,000 & 100,000 \\ \hline 
\# bc pts & 100,000 & 100,000 & 100,000 & N/A & 100,000 \\ \hline
$\lambda_{\text{res}}$ & 0.005 & 0.01 & 0.0001 & 0.1 & 0.01 \\
\bottomrule
\end{tabular}
\caption {Experimental details of the forward problems for training PIXELs: (96, 4, 16, 16) means, 96 multigrids, channel size of 4, the spatial grid size of 16, and temporal grid size of 16. \# collocation pts, \# ic pts, and \# bc pts denote the number of collocation, initial condition, and boundary condition points, respectively.}
\label{table:pixel_forward_details}
\end{table*}

\label{Inverse Problem}
Table~\ref{table:pixel_forward_details} and Table~\ref{table:pixel_inverse_details} shows the experimental details for the forward and inverse problems respectively. Note that all other hyperparameters of the forward problems and the inverse problems are same explained in main text, including the architecture of PINN, confirming that the proposed method is not sensitive to hyperparameters. For the inverse problems, The number of data points used for ground truth data points is 25,600 for convection equation, Burger equation, and Reaction-diffusion equation. The Helmholtz equation uses 490,000 and the Allen-Cahn equation uses 102,912.

\begin{table*}[t]
\centering
\begin{tabular}{l|c|c|c|c|c}
\toprule
& Convection & Reaction & Helmholtz & Allen-Cahn & Burgers \\ & & diffusion & ($a_1=1, a_2=4$) & & \\ \hline
Grid sizes &  (192, 4, 16, 16) &  (192, 4, 16, 16) &  (16, 4, 16, 16)& (192, 4, 16, 16) & (192, 4, 16, 16) \\ \hline 
\# collocation pts & 100,000 & 100,000 & 100,000 & 100,000 & 100,000 \\ \hline
\# ic pts & 100,000 & 100,000 & N/A & 100,000 & 100,000 \\ \hline 
\# bc pts & 100,000 & 100,000 & 100,000 & N/A & 100,000 \\ \hline
$\lambda_{\text{res}}$ & 0.005 & 0.005 & 0.00001 & 0.1 & 0.0005 \\
\bottomrule
\end{tabular}
\caption {Experimental details of the inverse problems (PIXEL)}
\label{table:pixel_inverse_details}
\end{table*}

\section{Grid size and the number of data points}
\begin{table*}[ht]
\centering
\small {
\begin{tabular}{l|c|c|c|c|c}
\toprule
Multigrid sizes  & 5,000 (\# pts) & 10,000 & 20,000 & 50,000 & 100,000 \\ \hline
(4, 4, 16, 16) & 2.35e-01 &  2.32e-01 &    2.25e-01     &     2.23e-01    &  2.21e-01           \\ \hline     
(8, 4, 16, 16) & 5.59e-02 &  4.11e-02 &    3.10e-02     &   2.95e-02      &   3.73e-02          \\ \hline   
(16, 4, 16, 16) & 4.47e-02 &  2.99e-02 &    2.51e-02     &   8.01e-03      &   1.91e-02          \\ \hline 
(32, 4, 16, 16) & 4.40e-02&  2.69e-02 &   1.13e-02      &   1.13e-02      &  8.22e-03         \\
\bottomrule
\end{tabular}
}
\caption {Varying the amounts of training points and the number of multigrids ($L_2$ relative errors)}
\label{table:num_datapoints}
\end{table*}

This section studies the relationship between the amounts of training data points and the grid sizes. We introduced multigrid representations that inject a smooth prior into the whole framework, which would reduce the required data points for each training iteration. We demonstrate this with the convection equation example by varying the number of training data points (collocation and initial condition) and the number of multigrid representations. We fixed the grid size 16 and the channel size 4, and varied the number of multigrids from 4 to 64. We reported the $L_2$ relative errors after 500 L-BFGS iterations. As shown in Table~\ref{table:num_datapoints}, our method is robust to the amounts of training data points. Although we can achieve higher accuracy with more training points, it performs comparably with a few data regimes. 

\section{The visualization of multigrid representations}
\begin{figure*}[ht]
    \centering
    \includegraphics[width=0.8\textwidth]{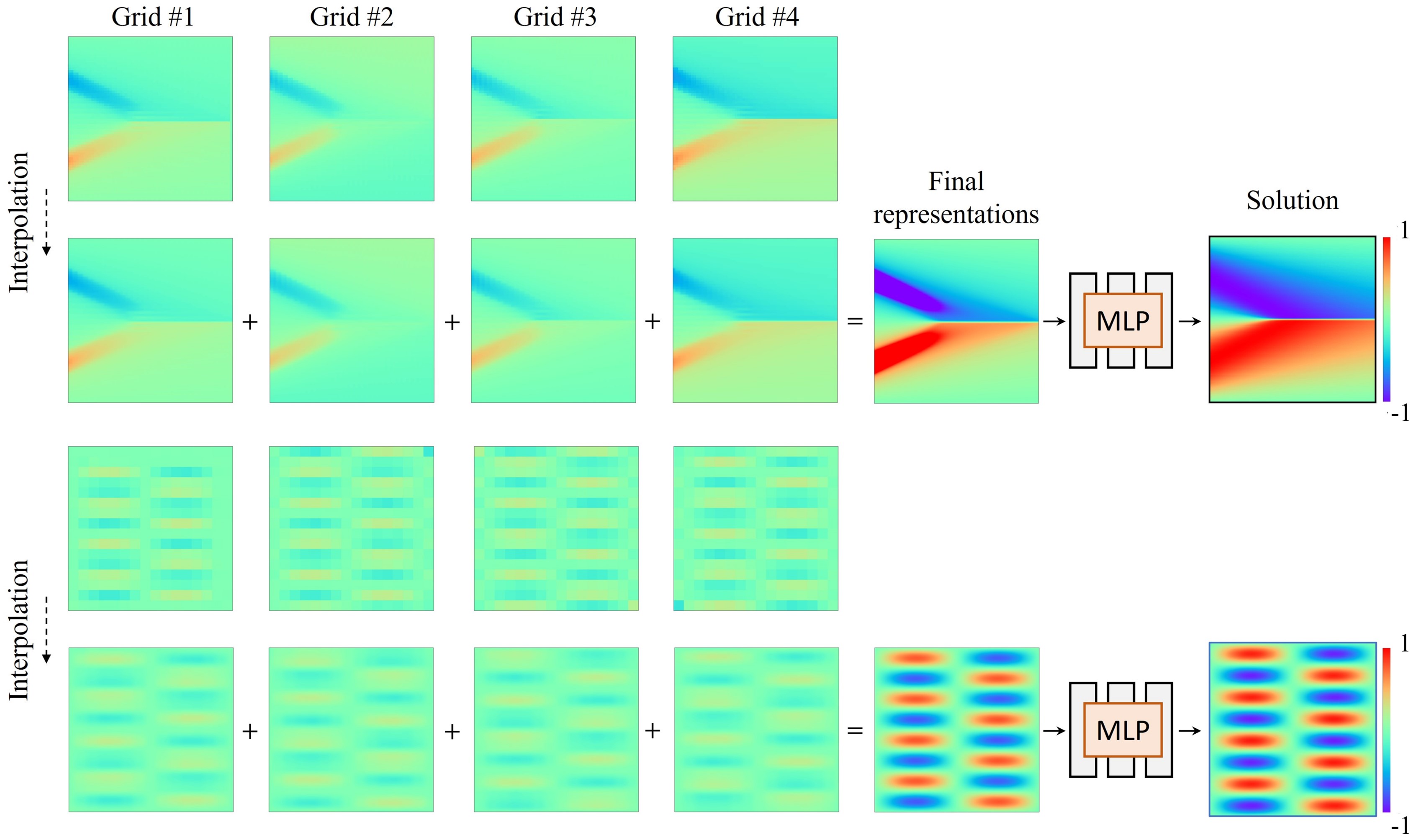}
    \caption{Visualization of multigrid representations for Burgers and Helmholtz equations (best viewed zoomed in): The first row shows image plots of each grid representation, and the second row shows the representations after the interpolation. The final representations are obtained through the sum of each interpolated cell, followed by an MLP to generate the solution. We used (4,4,64,64) and (4,4,16,16) multigrid representations for Burgers and Helmholtz, respectively.}
    \label{fig:multicell_viz}
\end{figure*}
We demonstrate the intermediate results in Figure~\ref{fig:multicell_viz}. In Burgers example (the first and the second rows), we used (4,1,64,64) configuration and two layers of MLP with 16 hidden units. As we expect, each grid show foggy images since the final solution will be the sum of all multigrid representations. Also, we shifted each grid, which resulted in slightly different images from each other. The final solution is completed in the last stage by filling up the remaining contents using an MLP. Importantly, we note that the singular behavior (shock, a thin line in the middle of the solution image) is already well captured by the grid representations. The role of MLP was to represent the smooth global component in solution function. Therefore, the proposed grid representations and an MLP combine each other's strengths to obtain better final solutions.

In Helmholtz example, we used small size grids (4,4,16,16). Thus, we can observe notable differences after cosine interpolation (grid-like pattern before the interpolation). We also note that the grid representations already represent complex patterns, and the last MLP stage also refined the solution by darkening the colors in the solution image.

\section{Data}
We used publicly available data from \cite{raissi2019physics}, which is the ground truh data for Burgers and Allen-Cahn equations. 



\begin{figure*}[ht]
    \centering
    \includegraphics[width=0.8\textwidth]{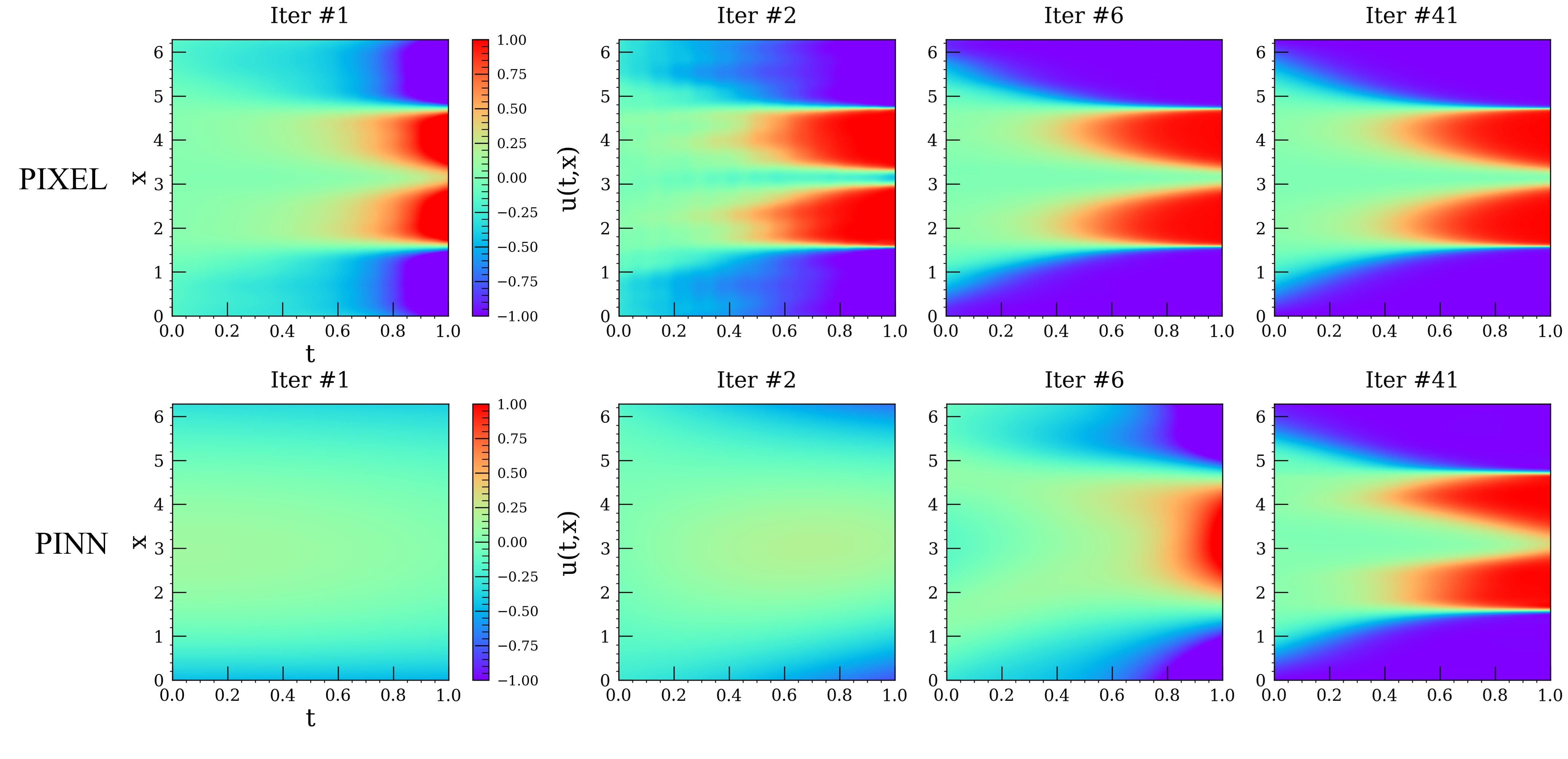}
    \caption{Inverse problem of allen-cahn equation}
    \label{fig:Allen-Cahn inverse plot}
\end{figure*}


\begin{figure*}
    \centering
    \includegraphics[width=0.8\textwidth]{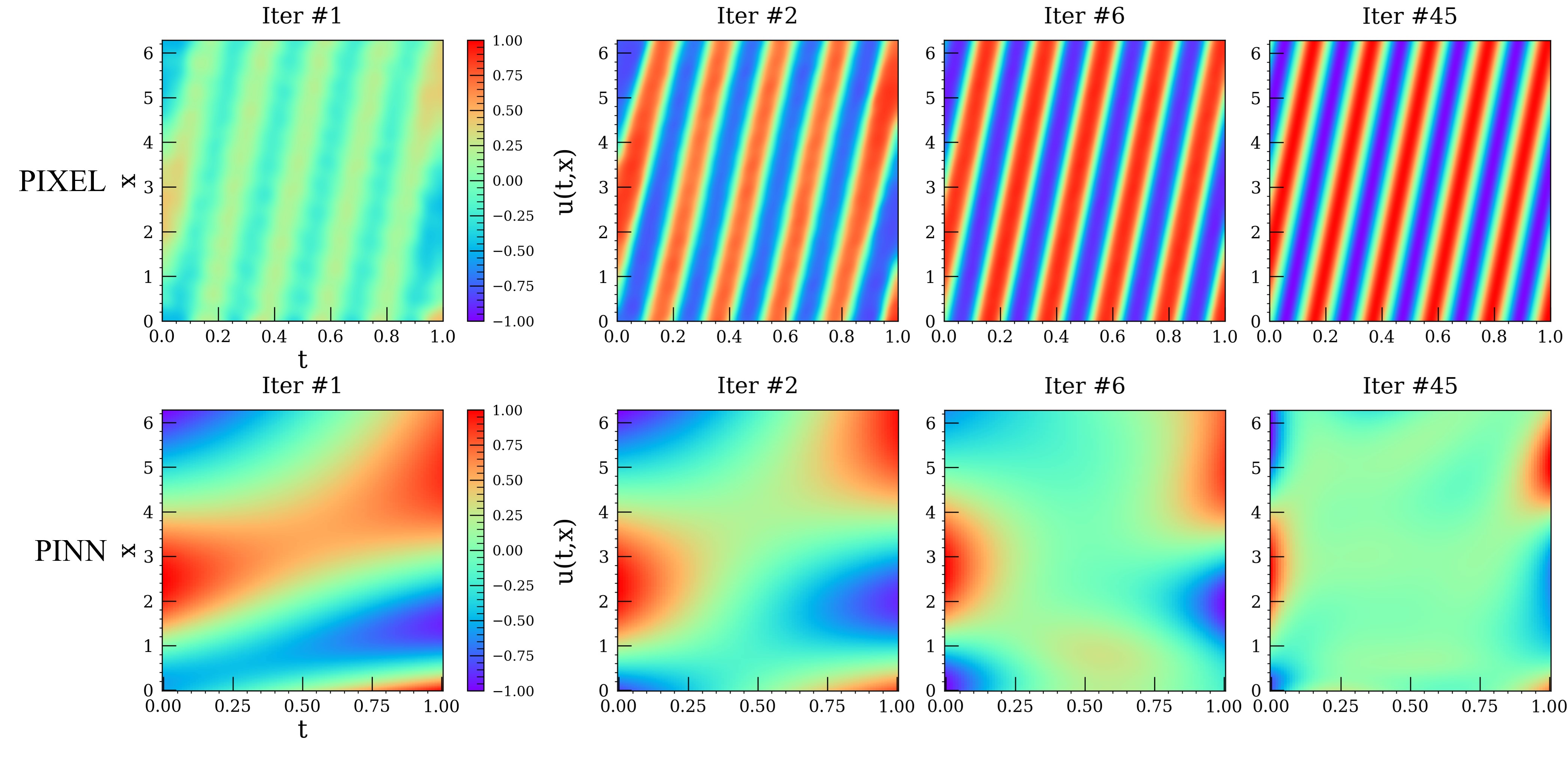}
    \caption{Inverse problem of convection equation}
    \label{fig:convection inverse result}
\end{figure*}



\begin{figure*}
    \centering
    \includegraphics[width=0.8\textwidth]{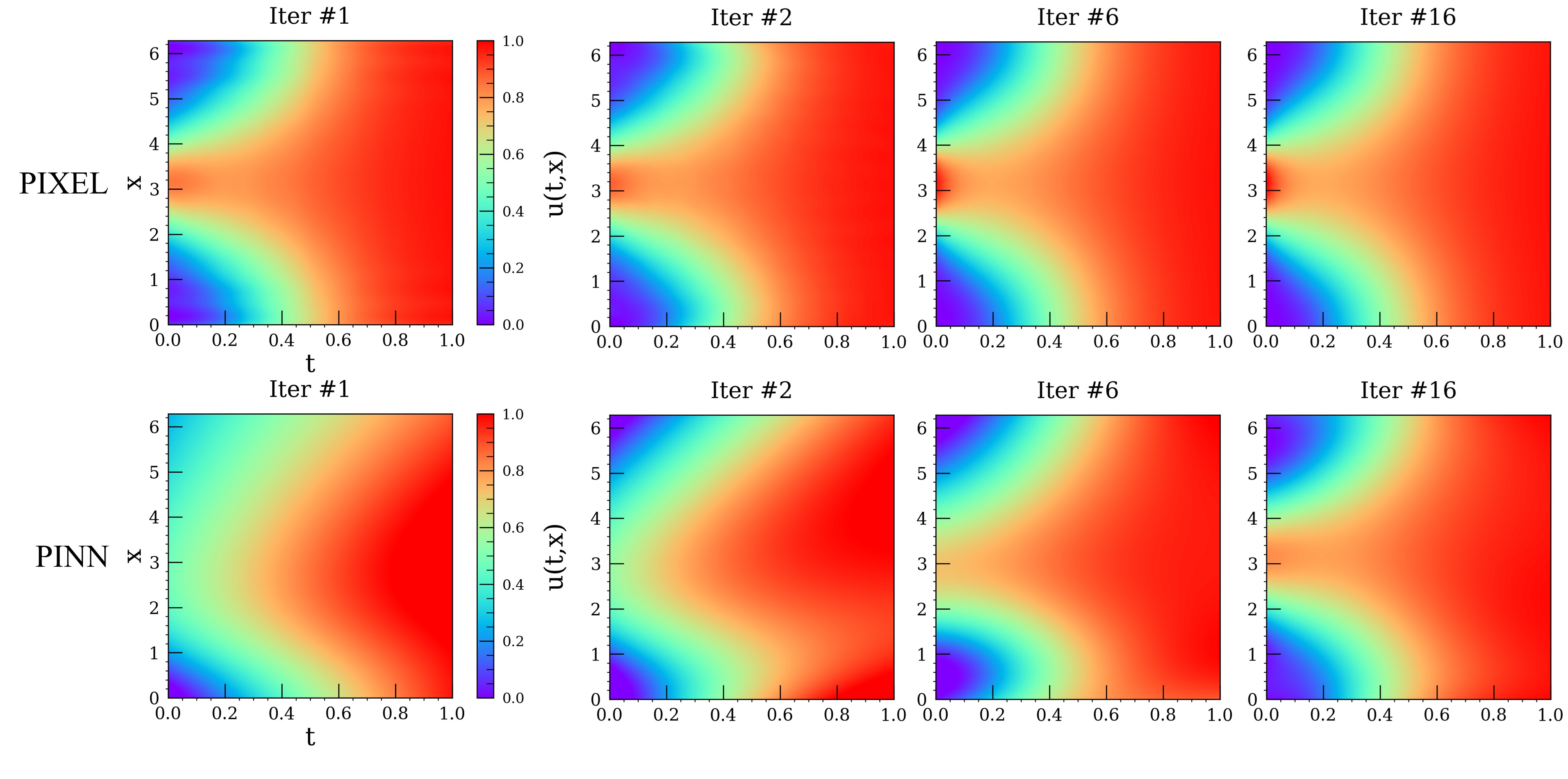}
    \caption{Inverse problem of reaction-diffusion equation}
    \label{fig:Reaction diffusion inverse}
\end{figure*}

\begin{figure*}
    \centering
    \includegraphics[width=0.8\textwidth]{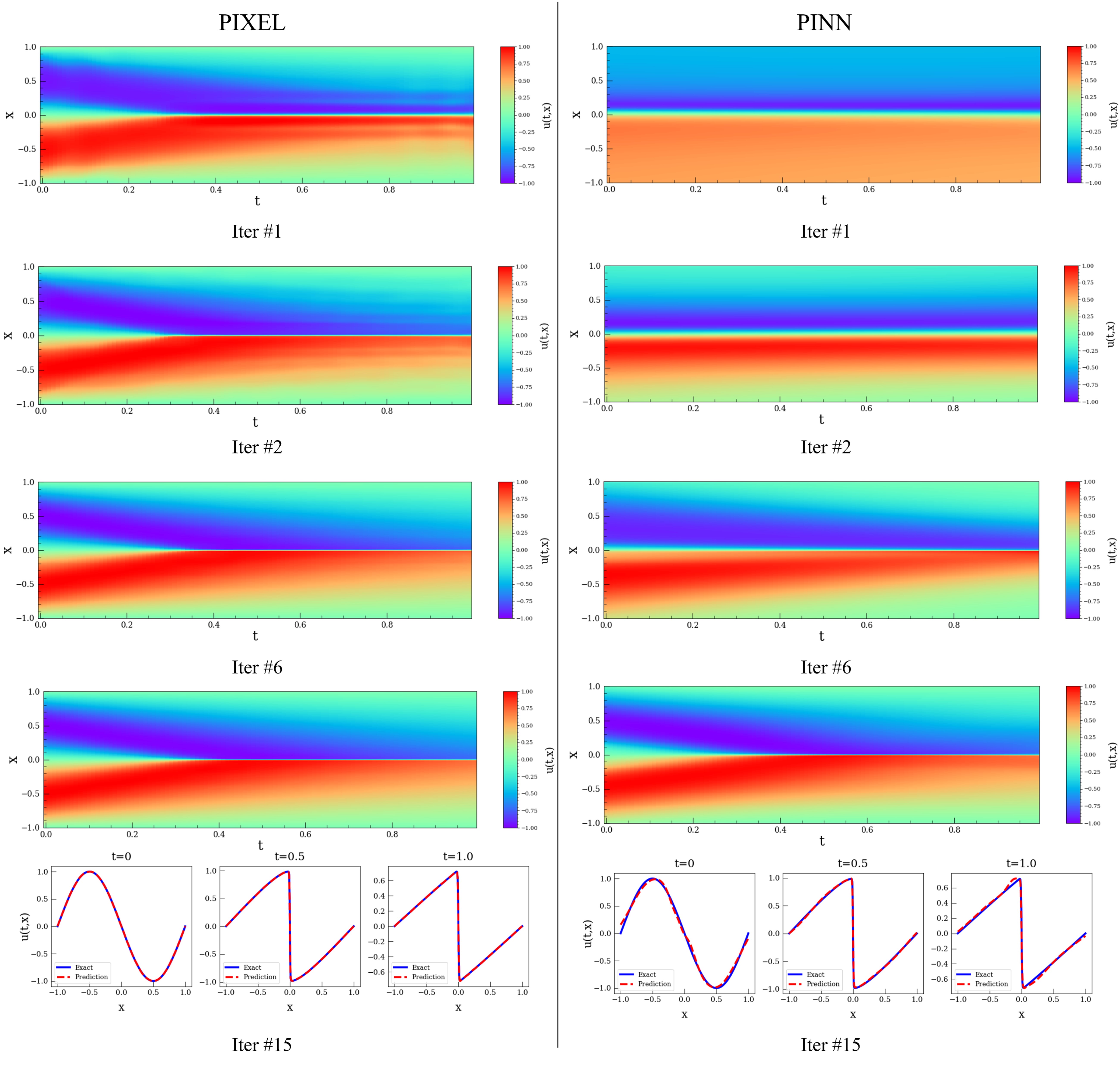}
    \caption{Inverse problem of burgers equation}
    \label{fig:Burgers inverse plot}
\end{figure*}



\begin{figure*}
    \centering
    \includegraphics[width=0.8\textwidth]{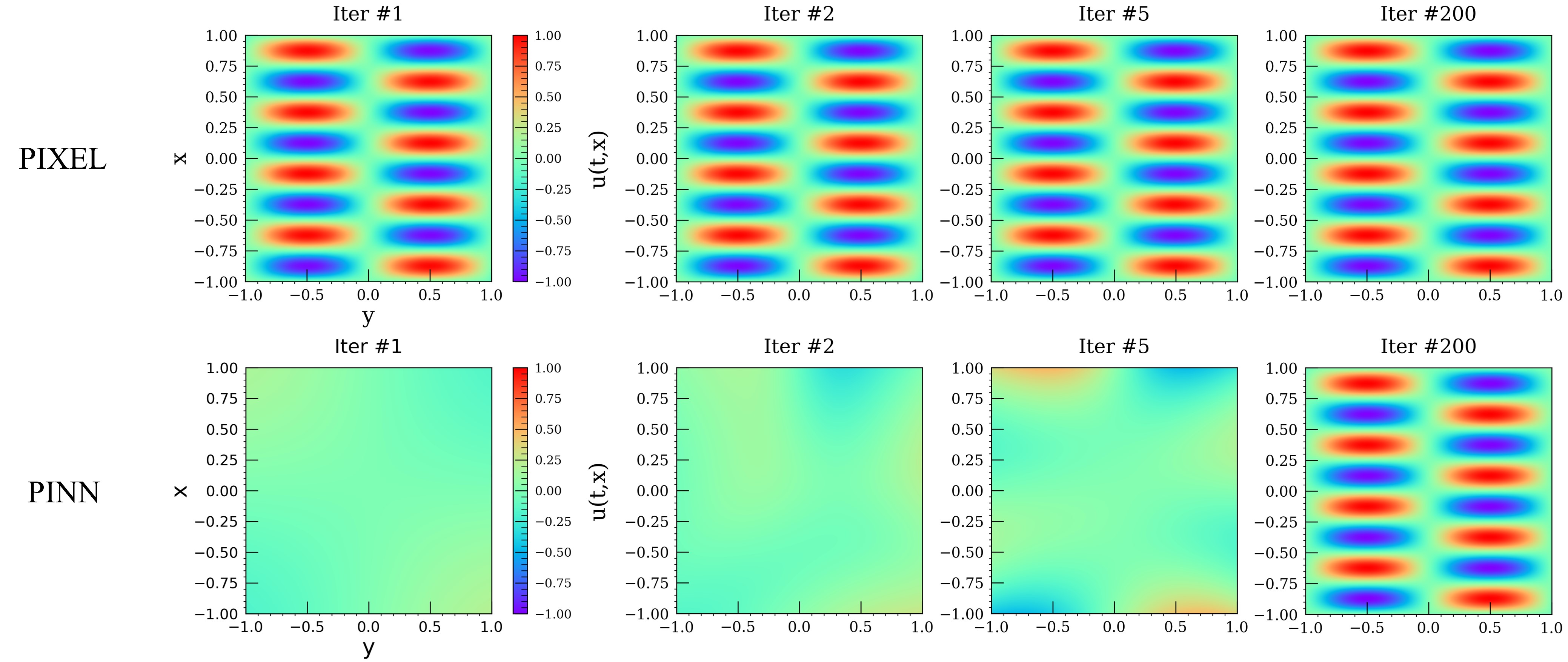}
    \caption{Inverse problem of helmholtz equation.}
    \label{fig:Helmholtz inverse plot}
\end{figure*}


\begin{figure*}
    \centering
    \includegraphics[width=0.8\textwidth]{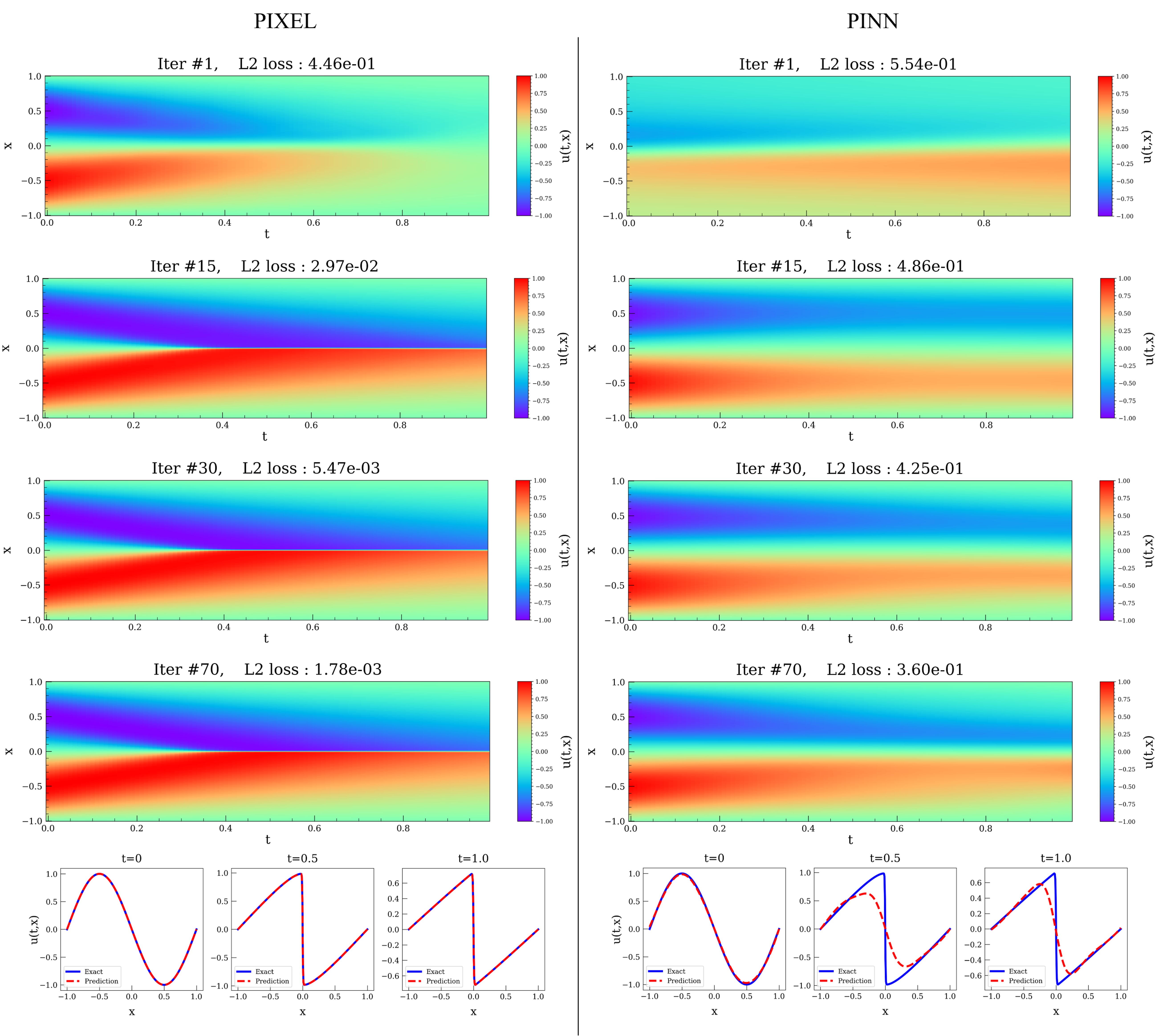}
    \caption{Forward problem of burgers equation}
    \label{fig:Burgers forward iter 1-70}
\end{figure*}



\begin{figure*}
    \centering
    \includegraphics[width=0.8\textwidth]{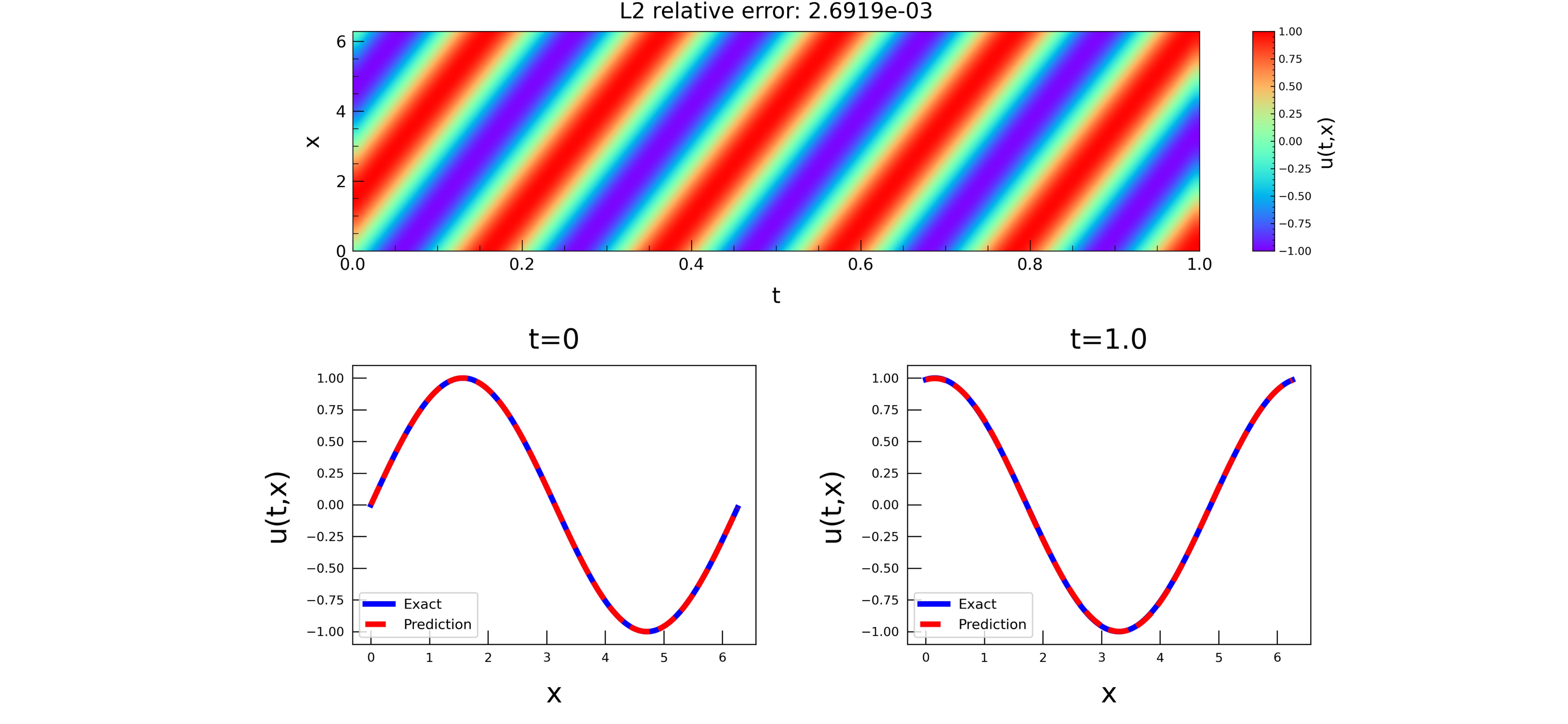}
    \caption{Convection equation results of PIXEL, at the 1500 iteration, the final relative $L_2$ error is 2.69e-03}
    \label{fig:Convection forward results}
\end{figure*}

\begin{figure*}
    \centering
    \includegraphics[width=0.8\textwidth]{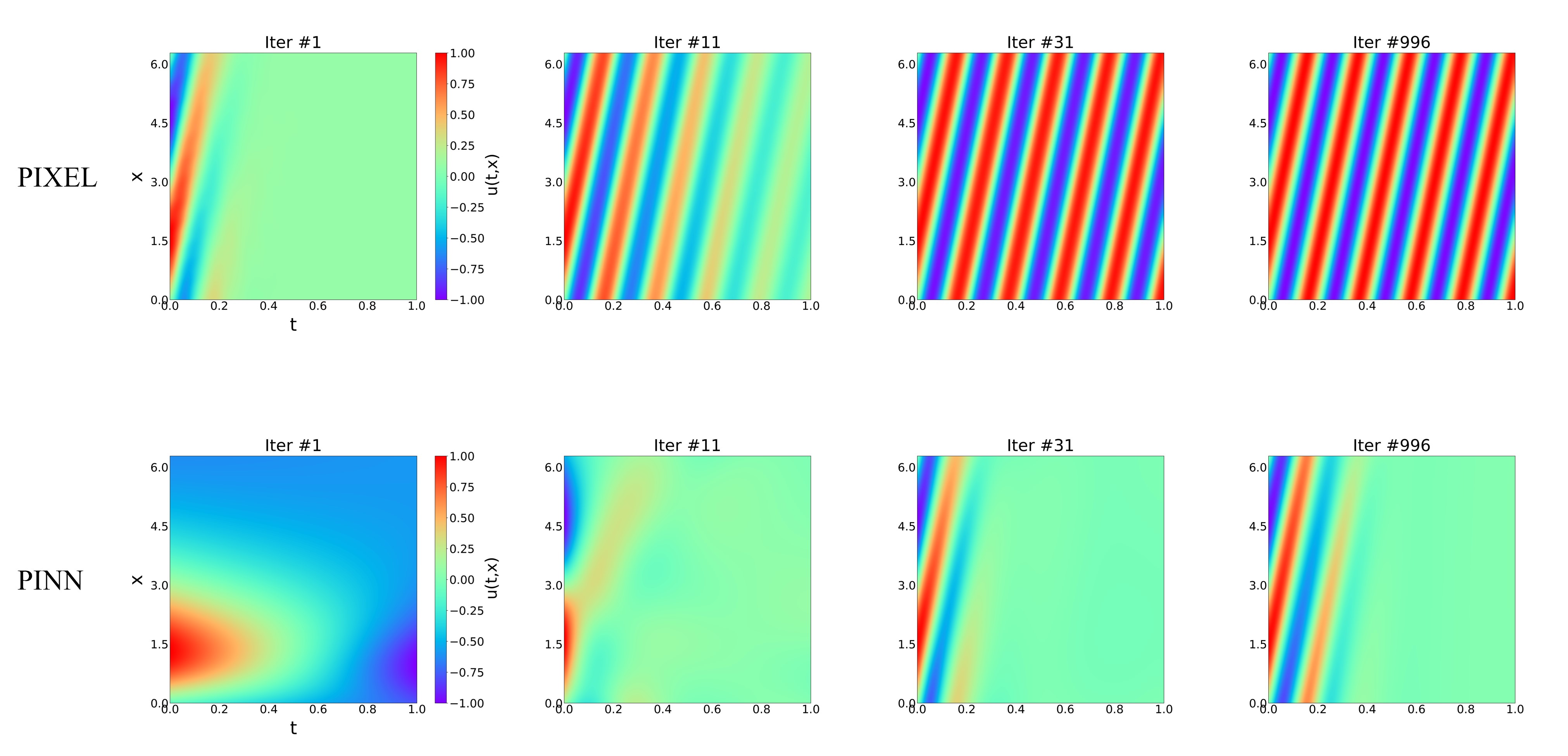}
    \caption{Forward problem of convection equation}
    \label{fig:Convection forward}
\end{figure*}

\begin{figure*}
    \centering
    \includegraphics[width=0.8\textwidth]{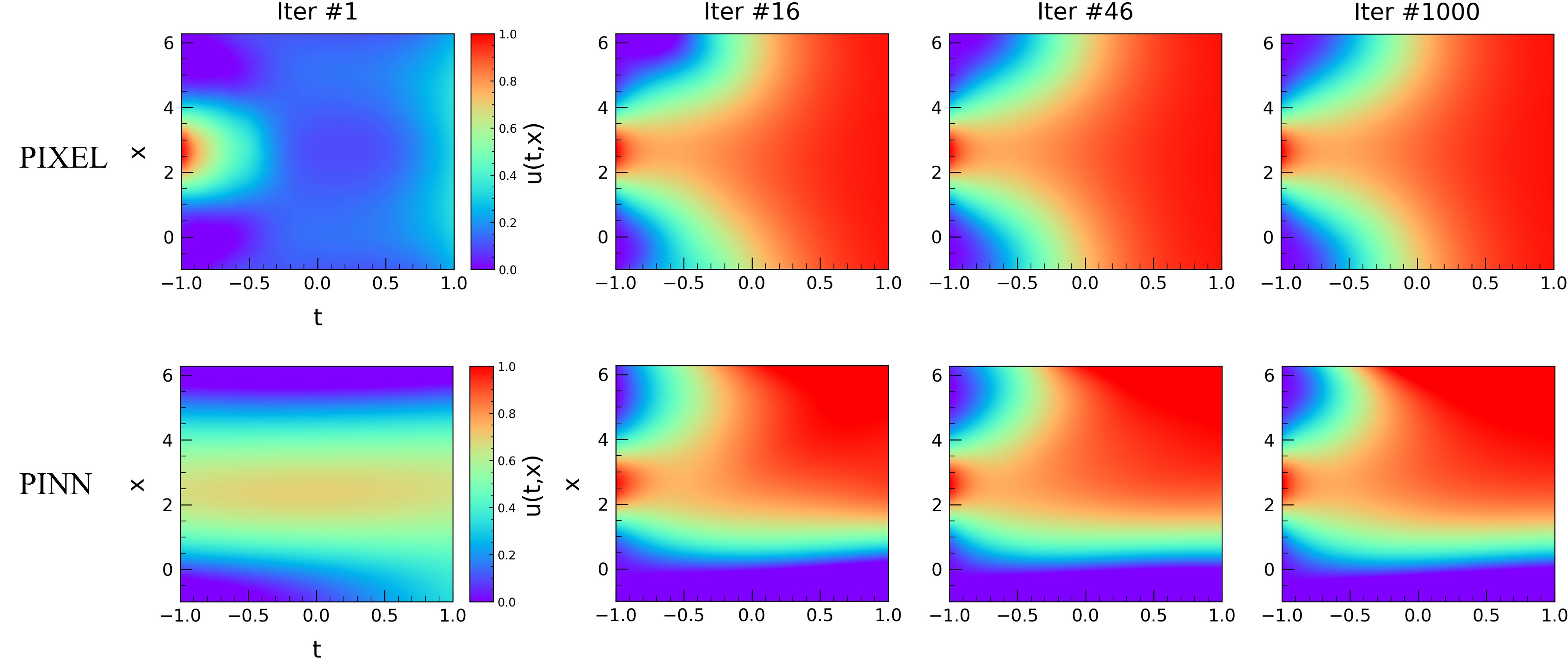}
    \caption{Forward problem of reaction-diffusion equation}
    \label{fig:Reaction diffusion forward}
\end{figure*}

\begin{figure*}
    \centering
    \includegraphics[width=0.75\textwidth]{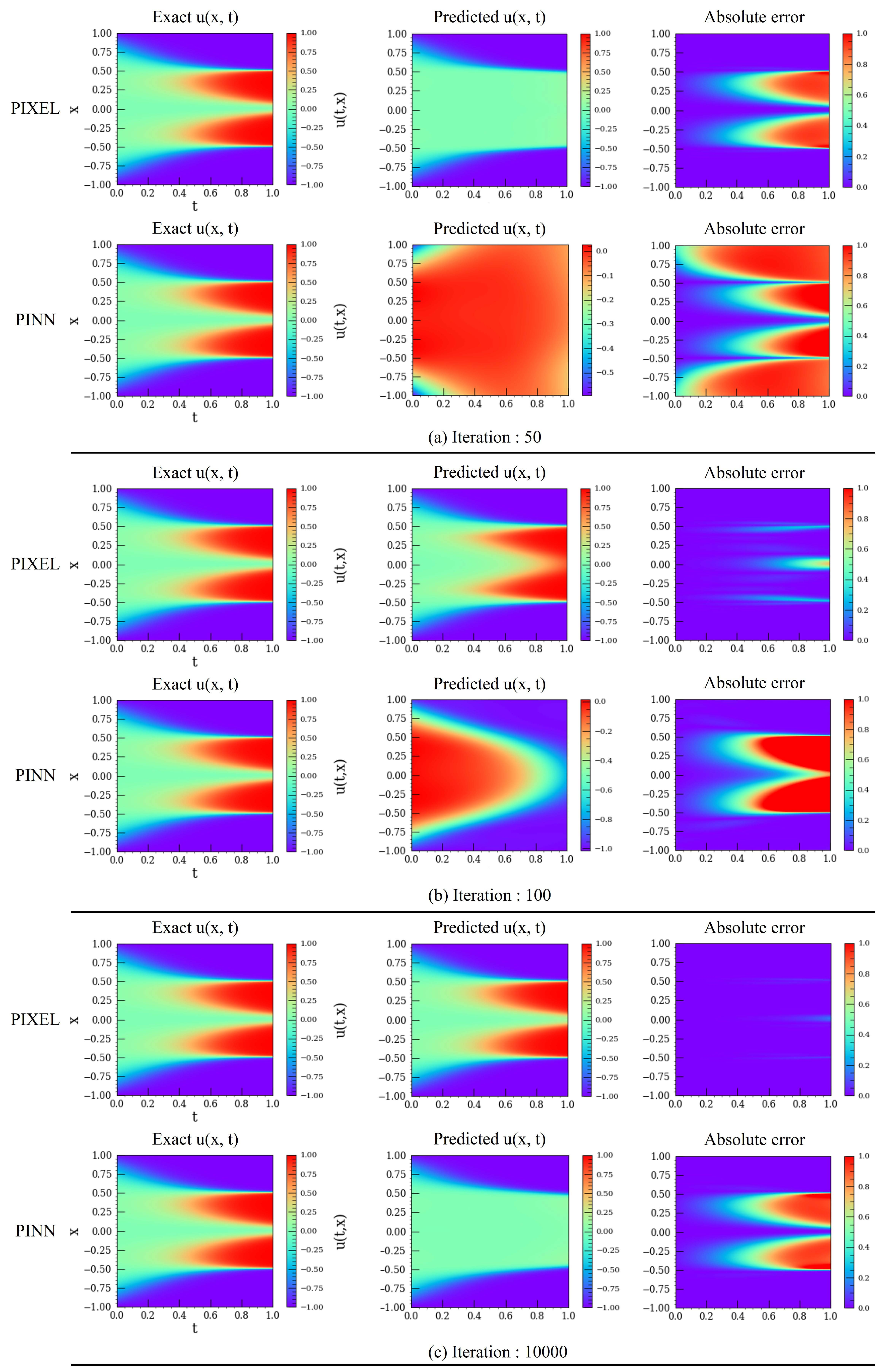}
    \caption{Forward problem of allen-cahn equation.}
    \label{fig:Allen-Cahn forward}
\end{figure*}

\begin{figure*}
    \centering
    \includegraphics[width=0.7\textwidth]{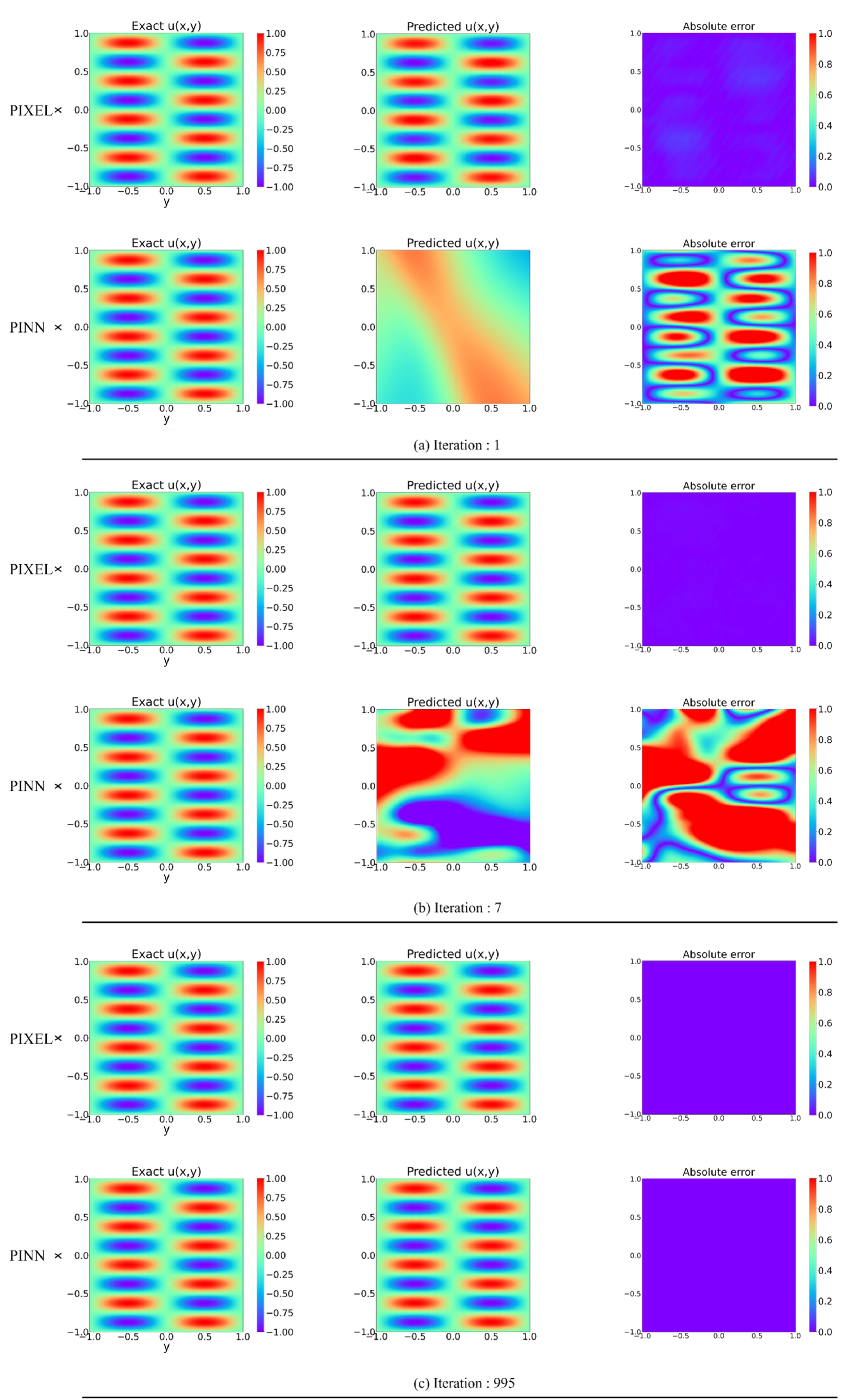}
    \caption{Forward problem of helmholtz equation}
    \label{fig:Helmholtz forward}
\end{figure*}

\bibliographynew{aaai23_}

\end{document}